\pdfoutput=1
\documentclass[11pt]{article}
\usepackage{natbib}
\usepackage{naacl2021}
\usepackage{times}
\usepackage{latexsym}
\usepackage[T1]{fontenc}
\usepackage[utf8]{inputenc}
\usepackage{microtype}
\usepackage{amsmath}
\usepackage{enumitem}
\usepackage{flushend}
\usepackage[section]{placeins}
\usepackage{todonotes}
\usepackage{tcolorbox}
\usepackage{multirow}
\usepackage{color,soul}
\RequirePackage{colortbl}
\definecolor{airforceblue}{rgb}{0.36, 0.54, 0.66}
\definecolor{amaranth}{rgb}{0.9, 0.17, 0.31}
\definecolor{applegreen}{rgb}{0.55, 0.71, 0.0}
\definecolor{alizarin}{rgb}{0.82, 0.1, 0.26}
\definecolor{azure}{rgb}{0.0, 0.5, 1.0}
\definecolor{cadmiumgreen}{rgb}{0.0, 0.42, 0.24}

\title{Challenges and Limitations with the Metrics Measuring the Complexity of Code-Mixed Text}

\author{Vivek Srivastava \\
  TCS Research\\ Pune, Maharashtra, India \\
  \texttt{srivastava.vivek2@tcs.com} \\\And
  Mayank Singh \\
  IIT Gandhinagar\\ Gandhinagar, Gujarat, India \\
  \texttt{singh.mayank@iitgn.ac.in} \\}

\begin{document}
\maketitle
\begin{abstract}
Code-mixing is a frequent communication style among multilingual speakers where they mix words and phrases from two different languages in the same utterance of text or speech. Identifying and filtering code-mixed text is a challenging task due to its co-existence with monolingual and noisy text. Over the years, several code-mixing metrics have been extensively used to identify and validate code-mixed text quality. This paper demonstrates several inherent limitations of code-mixing metrics with examples from the already existing datasets that are popularly used across various experiments. 
\end{abstract}

\section{Introduction}
\label{sec:intro}
Code-mixing is the phenomenon of mixing words and phrases from multiple languages in the same utterance of a text or speech \cite{bokamba1989there}. Multilingual societies observe a high frequency of code-mixed communication in the informal setting such as social media, online messaging, discussion forums, and online gaming~\cite{tay1989code}. Various studies indicate the overwhelming growth in the number of code-mixed speakers in various parts of the world, such as India, Spain, and China~\cite{baldauf2004hindi}. The phenomenal increase of the code-mixed data on various platforms such as Twitter, Facebook, WhatsApp, Reddit, and Quora, has led to several interesting research directions such as token-level language identification \cite{shekhar2020language,singh2018language}, POS tagging \cite{vyas2014pos,singh2018twitter}, machine translation \cite{dhar2018enabling,srivastava2020phinc}, and question-answering \cite{chandu2019code,banerjee2016first}.

Despite such active participation from the computational linguistic community in developing tools and resources for the code-mixed languages, we observe many challenges in processing the code-mixed data. One of the most compelling problems with the code-mixed data is the co-existence with the noisy and monolingual data. In contrast to the monolingual languages, we do not find any platform where the code-mixed language is the only medium of  communication. The co-existing nature of the code-mixed languages with the noisy and monolingual languages posits the fundamental challenge of filtering and identifying the code-mixed text relevant for a given study. Over the years, various works have employed human annotators for this task. However, employing humans for identifying and filtering the code-mixed text (in addition to the task-specific annotations) is extremely expensive on both fronts of time and cost. Also, since code-mixed languages do not follow specific linguistic rules and standards, it becomes increasingly challenging to evaluate human annotations and proficiency. 

\begin{table*}[!tbh]
\centering
\begin{tabular}{|cccc|}
\hline
\textbf{Data source} & \textbf{Task} & \textbf{Dataset size} & \textbf{Reported CMI} \\ \hline 
\newcite{singh2018named}   &      Named-entity recognition &3,638& Unavailable    \\ 
\newcite{swami2018corpus}&     Sarcasm detection&5,520&Unavailable \\ 
\newcite{joshi2016towards}&     Sentiment analysis & 3,879& Unavailable  \\ 
\newcite{patwa2020semeval}&     Sentiment analysis &    20,000     &   25.32\\ 
\newcite{barman-etal-2014-code}   &Language identification & 771&  13\\ 
\newcite{bohra2018dataset}&    Hate-speech detection  &      4,575   &  Unavailable\\ 
\newcite{dhar2018enabling}&     Machine translation &  6,096&     30.5\\ 
\newcite{srivastava2020phinc}&     Machine translation &13,738  &     75.76\\

\newcite{vijay2018dataset}&     Irony detection &3,055  &     Unavailable\\ 

\newcite{khanuja2020new}&     Natural language inference &2,240  &     >20\\ \hline

\end{tabular}
\caption{We explore 10 Hinglish code-mixed datasets to showcase the limitations of code-mixing metrics.}
\label{tab:datasets}
\end{table*}

In order to address some of the above challenges, several code-mixing metrics~\cite{das2014identifying,gamback2016comparing,doi:10.1177/13670069000040020101,guzman2017metrics} have been proposed to measure the degree of code-mixing in the text. However, we observe several limitations in the metric formulations. This paper outlines several such limitations and supports our claims with examples from multiple already existing datasets for various tasks. For illustrations, we choose Hinglish (code-mixing of Hindi and English language) due to two major reasons: (i) popularity of Hinglish and (ii) active research community. \citet{baldauf2004hindi} projected that number of Hinglish speakers might soon outrun the number of native English speakers in the world. This strengthens our belief that even though Hinglish (and other code-mixed languages) does not enjoy the official status, we need to build robust systems to serve the multilingual societies. With the availability of datasets and tools for the Hinglish language, we seek a boom in the active participation from the computational linguistic community to address various challenges.

\noindent \textit{Outline of the paper:} We formally define Hindi-English code-mixing in Section~\ref{sec:hinglish}. Section~\ref{sec:CM_metrics} describes several code-mixing metrics. We outline various limitations supported with multiple examples from various datasets in Section \ref{sec:limitations}. We conclude and present future direction in Section \ref{sec:conclusion}.

\section{Hinglish: Mixing Hindi with English}
\label{sec:hinglish}
Hinglish is a portmanteau of Hindi and the English language. Figure \ref{fig:cm_example} shows example Hinglish sentences. Also, we see two example sentences in Figure \ref{fig:cm_example} that are non-code-mixed but might appear to contain words from two languages. The presence of named entities from the Hindi language does not make the sentence code-mixed. 

\begin{figure}[!tbh]
\centering
\small{
\begin{tcolorbox}[colback=white]
\begin{center}
    \textbf{Code-mixed sentences}
\end{center}
\textsc{Sentence 1}: \textcolor{orange}{ye ek} \textcolor{blue}{code mixed sentence} \textcolor{orange}{ka} \textcolor{blue}{example} \textcolor{orange}{hai} \\
\textsc{Sentence 2} : \textcolor{orange}{kal me} \textcolor{blue}{movie} \textcolor{orange}{dekhne ja raha hu}. \textcolor{blue}{How are the reviews}? \\
\begin{center}
    \textbf{Non-code-mixed sentences}
\end{center}
\textsc{Sentence 1}: \textcolor{purple}{Tendulkar} \textcolor{blue}{scored more centuries than} \textcolor{purple}{Kohli} \textcolor{blue}{in} \textcolor{purple}{Delhi}.\\
\textsc{Sentence 2}: \textcolor{purple}{Bhartiya Janta Party} \textcolor{blue}{won the 2019 general elections}.  \\
\end{tcolorbox}}
\caption{Example code-mixed sentences with words from \textcolor{orange}{Hindi} and \textcolor{blue}{English} languages. The non-code-mixed sentences might get confused with the code-mixed sentence due to the presence of \textcolor{purple}{named entities}.}
\label{fig:cm_example}
\end{figure}

In this study, we explore 10 Hinglish datasets encompassing eight different tasks, namely named entity recognition, sarcasm detection, sentiment analysis, language identification, hate-speech detection, machine translation, irony detection, and natural language inference (see Table~\ref{tab:datasets} for more details). Contrasting against monolingual datasets for similar tasks, the Hinglish datasets are significantly smaller in size. We support our claims by providing illustrative examples from these datasets.

\section{Code-Mixing Metrics}
\label{sec:CM_metrics}
In this section, we describe several popular code-mixing metrics that measure the complexity of the code-mixed text. Among the following metrics, code-mixing index (CMI,~\citet{das2014identifying,gamback2016comparing}) is the most popular metric. 

\subsection{Code-mixing Index}
\label{sec:CMI}

CMI metric~\cite{das2014identifying} is defined as follows:
\begin{equation}
\label{eq:oldCMI}
    CMI= \begin{cases} 
100 * [1- \frac{max(w_{i})}{n-u}] & n> u \\
0 & n=u 
\end{cases}
\end{equation} 

Here, $w_{i}$ is the number of words of the language $i$, max\{{$w_{i}$}\} represents the number of words of the most prominent language, $n$ is the total number of tokens, $u$ represents the number of language-independent tokens (such as named entities, abbreviations, mentions, and hashtags).

A low CMI score indicates monolingualism in the text whereas the high CMI score is an indicator of the high degree of code-mixing in the text. In the later work,~\cite{gamback2016comparing} also introduced number of code  alternation points in the original CMI formulation. An \textit{alternation point} (a.k.a. \textit{switch point}) is defined as any token in the text that is preceded by a token with a different language tag. Let $f_p$ denotes ratio of  number of code alternation points $P$ per token, $f_p = \frac{P}{n}$ where $0 \le P <n$. Let CMI$_{old}$ denotes the CMI formulation defined in Eq.~\ref{eq:oldCMI}. The updated CMI formulation (CMI$_{new}$) is defined as:
\begin{equation}
\label{eq:newCMI}
    CMI_{new}=  a.CMI_{old} + b.f_p
\end{equation} 

where $a$ and $b$ are   weights, such that $a+b =1$. Again, $CMI_{new}$ = 0 for monolingual text, as  $CMI_{old}$ = 0 and $P=0$. Hereafter, throughout the paper, we refer to $CMI_{new}$ as CMI metric. 


\subsection{M-index}
\label{sec:Mindex}
\citet{doi:10.1177/13670069000040020101} proposed the Multilingual Index (M-index). M-index measures the inequality of the distribution of language tags in a text comprising at least two languages. If $p_j$ is the total number of words in the language $j$ over the total number of words in the text, and $j \in k$, where k is total number of languages in the text, M-index is defined as:
\begin{equation}
    M-index= \frac{1-\sum p_{j}^2 }{(k-1)\sum p_{j}^2}
\end{equation}

The index varies between 0 (monolingual utterance) and 1 (a perfect code-mixed text comprising equal contribution from each language).

\subsection{I-index}
\label{sec:Iindex}
The Integration-index proposed by \citet{guzman2017metrics} measures the probability of switching within a text. I-index approximates the probability that any given token in the corpus is a switch point. Consider a text comprised of $n$ tokens, I-index is defined as:

\begin{equation}
    I-index= \frac{\sum_{1\leq i <n-1} S(i,i+1)}{n-1}
\end{equation}

Here, $S(i,i+1) = 1$ if language tag of $i^{th}$ token is different than the language tag of ${(i+1)}^{th}$ token, otherwise $S(i,i+1) = 0$. I-index varies between 0 (monolingual utterance) and 1 (a perfect code-mixed text comprising consecutive tokens with different language tag). \citet{guzman2017metrics} also adapted two metrics that quantify burstiness and memory in complex systems~\cite{goh2008burstiness} to measure the complexity of code-mixed text. Next, we introduce these complex system-based metrics. 

\subsection{Burstiness}
\label{sec:burstiness}
Burstiness~\cite{goh2008burstiness} measures whether switching occurs in bursts or has a more periodic character. Let $\sigma_r$ denote the standard deviation of the language spans and $m_r$ the mean of the language spans. Burstiness is calculated as:

\begin{equation}
    Burstiness = \frac{\sigma_r - m_r}{\sigma_r + m_r}
\end{equation}

The burstiness metric is bounded within the interval [-1, 1]. Text with periodic dispersions of switch points yields a burstiness value closer to -1.  In contrast, text with high burstiness and containing less predictable switching patterns take values closer to 1. 

\subsection{Memory}
\label{sec:memory}
Memory~\cite{goh2008burstiness} quantifies the extent to which the length of language spans tends to be influenced by the length of spans preceding them. Let $n_r$ be the number of language spans in the utterance and $\tau_i$ denote a specific language span in that utterance ordered by $i$.  Let $\sigma_1$ and $\mu_1$ be the standard deviation and mean of all language spans but the last,  where $\sigma_2$ and $\mu_2$ are the standard deviation and mean of all language spans but the first.

\begin{equation}
    Memory = \frac{1}{n_r -1} \sum^{n_r -1}_{1} \frac{(\tau_i - \mu_1)(\tau_{i+1} - \mu_2)}{\sigma_1 \sigma_2}
\end{equation}

Memory varies in an interval [-1,1]. Memory values close to -1 describe the tendency for consecutive language spans to be negatively correlated, that is, short spans follow long spans,  and vice-versa. Conversely, memory values closer to 1 describe the tendency for consecutive language spans to be positively correlated, meaning similar in length.

In addition to the above metrics, there exist several other code-mixing metrics such as Language Entropy and Span Entropy that can be derived from the above metrics~\cite{guzman2017metrics}. Due to the space constraints, we refrain from further discussing them in the paper. 

\begin{table*}[!tbh]
\resizebox{\hsize}{!}{
\begin{tabular}{|c|c|c|c|c|c|c|c|c|c|c|c|}
\hline
\multirow{2}{*}{\textbf{Hinglish sentence}} & \multirow{2}{*}{\textbf{CMI}} & \multirow{2}{*}{\textbf{M-index}} & \multirow{2}{*}{\textbf{I-index}} & \multirow{2}{*}{\textbf{Burstiness}} & \multirow{2}{*}{\textbf{Memory}} & \multicolumn{2}{c|}{\textbf{Human 1}} & \multicolumn{2}{c|}{\textbf{Human 2}} & \multicolumn{2}{c|}{\textbf{Human 3}} \\ \cline{7-12} 
  &                      &                          &                          &                             &                         & 
 \textbf{DCM}       & \textbf{RA}      & \textbf{DCM}       & \textbf{RA}  & \textbf{DCM}       & \textbf{RA}      \\ \hline
 
 \begin{tabular}[c]{@{}c@{}}Deepak \textcolor{orange}{ji}, \textcolor{blue}{channel} \textcolor{orange}{ko kitna} \textcolor{blue}{fund} \textcolor{orange}{diya} \\\textcolor{orange}{hai} congress \textcolor{orange}{ne}? 2006 \textcolor{orange}{me} ameithi \textcolor{blue}{rape}\\\textcolor{blue}{case} \textcolor{orange}{kyu nahi} \textcolor{blue}{discuss} \textcolor{orange}{kiya kabhi}?\end{tabular}   &   3.53                   &             7.59             &    7.27                      &        -0.46       &        -0.12      &           8         & 10            & 9         & 10               &      10        & 8            \\ \hline
    
  \begin{tabular}[c]{@{}c@{}}4 \textcolor{orange}{din me} 2 \textcolor{blue}{accidents}, \textcolor{orange}{kuch to jhol}\\  \textcolor{orange}{hai}, \textcolor{orange}{shayad} \textcolor{blue}{politics} \textcolor{orange}{ho rahi hai}..\end{tabular}  &       1.67               &         6.2                 &              5           &         -0.19                    &             -0.31            &               4         & 9             & 5         & 10               & 10        & 9                  \\ \hline
    
  \begin{tabular}[c]{@{}c@{}}\textcolor{orange}{Bhai kasam se bata do ki shadi kab karr}\\  \textcolor{orange}{rahe ho warna mai kuwara marr jaunga}\end{tabular} &       0               &         0                 &         0                 &      -1                       &          -0.41               &                0         & 10            & 1         & 10               & 9         & 7                  \\ \hline
    
  \begin{tabular}[c]{@{}c@{}}@Mariam\_Jamali \textcolor{blue}{Nice one but logo}\\ \textcolor{orange}{filhal} KK \textcolor{orange}{ki jaga} Pakistan \textcolor{orange}{ka lagwa}\\ \textcolor{orange}{do}. \textcolor{blue}{Pic is good}\end{tabular}  &       4.6               &                 9.7         &          4.7                &           -0.28                  &             -0.37            &              6         & 6              & 8         & 8                & 7         & 9                  \\ \hline
    
 \begin{tabular}[c]{@{}c@{}}\textcolor{orange}{abe} .,., \textcolor{blue}{joke} \textcolor{orange}{marna hai hi to aur hi kahi}\\ \textcolor{orange}{maar} .,.,. \textcolor{blue}{confession page} \textcolor{orange}{ki bejaati maat}\\ \textcolor{orange}{ker bhai} .. \textcolor{blue}{JOKE} \textcolor{orange}{MARA}???????????? \\\textcolor{orange}{HASU}? \textbackslash{}"haha..!\textbackslash{}"\end{tabular}    &            2          &           6.67               &                4.28          &                 -0.08            &      -0.18                   &     6         & 5                & 3         & 8                & 7         & 5                  \\ \hline
    
   \begin{tabular}[c]{@{}c@{}}\textcolor{orange}{Wale log jante hai par atankwadiyo}\\ \textcolor{orange}{nafrat failane walo ke liye meri} \\\textcolor{orange}{yehi} \textcolor{blue}{language} \textcolor{orange}{rahegi}\end{tabular} &     0.6                 &              1.42            &         1.42        &    0.09       &         0         &    4         & 6                      & 2         & 8                &  7         & 6         \\ \hline
    
  \begin{tabular}[c]{@{}c@{}}\textcolor{orange}{mujhe hasi aa rahi thi} , \textcolor{blue}{while I ws} \\ \textcolor{blue}{reading them} . :P\end{tabular}  &    5                  &           9.32               &      2.5                    &       -0.24                      &             -0.64            &                   10        & 10       & 9         & 10               & 6         & 6                 \\ \hline
    
    \begin{tabular}[c]{@{}c@{}}\textcolor{blue}{laufed}  ... \textcolor{blue}{first u hav to correct ur english} \\ \textcolor{orange}{baad me sochna use} !!!\end{tabular} &            3.33          &             6.67             &          3.07                &                  0.2          &          -0.06               &            10        & 8                & 8         & 9                & 6         & 7               \\ \hline
    
    \begin{tabular}[c]{@{}c@{}}\textcolor{blue}{The ultimate twist} \textcolor{orange}{Dulhan dandanate huye} \\ \textcolor{blue}{brings} \textcolor{orange}{Baraat} ....  \textcolor{orange}{Dulha}\end{tabular} &    4.44                  &              6.9            &        5.55                  &         -0.08      &       0.48       &     8         & 6              & 7         & 2                & 5         & 7               \\ \hline
    
   \begin{tabular}[c]{@{}c@{}}RAHUL \textcolor{orange}{jab} \textcolor{blue}{dieting} \textcolor{orange}{par hota hai toh} \\ \textcolor{blue}{green tea} \textcolor{orange}{peeta hai}.\end{tabular} &    3.63                  &          6.61                &      5.45                    &         -0.44                    &              0           &    10        & 10                 & 2         & 10               & 10        & 9               \\ \hline
\end{tabular}}
\caption{Measuring the complexity of various \textcolor{orange}{Hindi}-\textcolor{blue}{English} code-mixed text. Language independent tokens are marked with black color. We select one sentence each from the 10 datasets (in the same order as given in Table \ref{tab:datasets}). Here, DCM stands for degree of code-mixing and RA stands for readability. We scale the CMI, M-index, and I-index metric scores in the range 0 to 10. The range for Burstiness and Memory score is -1 to 1.}
\label{tab:metric_score}
\end{table*}

\noindent{\textbf{Evaluating metric scores on code-mixed datasets}:} To understand the effectiveness of these metrics, we randomly sample one sentence each from the ten datasets and calculate the score on all the code-mixing metrics. In addition, we employ three human annotators proficient in both the languages (English and Hindi) to rate the sentences on two parameters: the degree of code-mixing and readability. We provide the following guidelines to the annotators for this task:
\begin{itemize}
    \item \textbf{Degree of code-mixing (DCM)}: The score can vary between 0 to 10. A DCM score of 0 corresponds to the monolingual sentence with no code-mixing, whereas the DCM score of 10 suggests the high degree of code-mixing.
    \item \textbf{Readability (RA)}: RA score can vary between 0 to 10. A completely unreadable sentence due to large number of spelling mistakes, no sentence structuring, or meaning, yields a RA score of 0. A RA score of 10 suggests a highly readable sentence with clear semantics and easy-to-read words.
\end{itemize}

Table \ref{tab:metric_score} shows the 10 example Hinglish sentences with the corresponding metric scores and the human evaluation. Some major observations are:
\begin{itemize}
    \item We do not observe any metric to independently measure the readability of code-mixed text as quantified by humans.
    \item We also observe contrasting scores given by different metrics, making it difficult to choose the best-suited metric for the given code-mixed dataset. 
    \item At times, we observe a high disagreement even among the human ratings. This behavior indicates the complexity of the task for humans as well.
    \item We do not observe any significant relationship between the degree of code-mixing and the readability score as provided by humans. This observation is critical in building high-quality datasets for various code-mixing tasks.
\end{itemize}


\section{Limitations of code-mixing metrics}
\label{sec:limitations}
This section describes various limitations of the existing metrics that measure the complexity of the code-mixed text. As CMI is most popular among code-mixing metrics, it is reported in five \cite{patwa2020semeval,barman-etal-2014-code,dhar2018enabling,srivastava2020phinc,khanuja2020new} out of the 10 datasets listed in Table~\ref{tab:datasets}.  We describe major limitations of code-mixing metrics from three different perspectives:

\begin{table*}[!tbh]
\centering
\resizebox{\hsize}{!}{
\begin{tabular}{|c|c|c|c|}
\hline
\textbf{Data source} & \textbf{\textcolor{magenta}{Spelling variations}} & \textbf{\textcolor{blue}{Noisy}/\textcolor{orange}{monolingual}} & \textbf{\textcolor{blue}{Readability}/\textcolor{purple}{semantic}} \\ \hline
\newcite{singh2018named} &    \begin{tabular}[c]{@{}c@{}}   Ab boliye teen talak \textcolor{magenta}{harram}\\ \textcolor{magenta}{h} ya \textcolor{magenta}{nai} aapke khud ki lady's\\ \textcolor{magenta}{chate h} ki aap \textcolor{magenta}{sai dur} hona.\\ Shame on \textcolor{magenta}{u} again...\#TripleTalaq    \end{tabular} &\textcolor{orange}{\#TripleTalaq Don't post this}      &  \begin{tabular}[c]{@{}c@{}}   \textcolor{purple}{@BJP4UP @narendramodi} \\ \textcolor{purple}{@AmitShah @BJPLive @bjpsamvad}\\ \textcolor{purple}{@BJP4India \#NoteBandi ke baad}\\ \textcolor{purple}{ab poori}  \end{tabular}  \\ \hline

\newcite{swami2018corpus} &\begin{tabular}[c]{@{}c@{}}Shareef wo hai \textcolor{magenta}{jisay}\\ \textcolor{magenta}{moqa nae} milta! \#irony \end{tabular}    &\begin{tabular}[c]{@{}c@{}} \textcolor{blue}{Resigned: Sri Lanka Cricket aniyin} \\ \textcolor{blue}{thodar thoalviyinaal Therivuk kuluth}\\ \textcolor{blue}{Thalaivar Sanath Jayasuriya ullitta}\\ \textcolor{blue}{athan uruppinarhal Raajinaamaa}     \end{tabular} &     \begin{tabular}[c]{@{}c@{}}\textcolor{purple}{Kudakudhinge dhuvasthamee?} \\ \textcolor{purple}{\#Maldives \#Politics} \end{tabular} \\ \hline

\newcite{joshi2016towards} &    \begin{tabular}[c]{@{}c@{}}\textcolor{magenta}{Nhi} ye log \textcolor{magenta}{apny lie ayeen} change \\karty he ye konsa mulk k \textcolor{magenta}{lie} \\\textcolor{magenta}{sochty} he har koi \textcolor{magenta}{apny lie} aur \textcolor{magenta}{apny} \\families k \textcolor{magenta}{lie} politics me\\ \textcolor{magenta}{he sary} chor \textcolor{magenta}{he}    \end{tabular}&  \begin{tabular}[c]{@{}c@{}}  \textcolor{blue}{\#Cricket News 6 Saal Team Ki}\\ \textcolor{blue}{Qeyadat Karna Mare Liye Izaz}\\ \textcolor{blue}{Hai Ab Kisi Our Ko Aagay}\\ \textcolor{blue}{Aana Chahiye Sabiq Captain} \\ \textcolor{blue}{AB De Villiers}     \end{tabular}     &   \textcolor{purple}{Hiii kam chhe}    \\ \hline

\newcite{patwa2020semeval} &    \begin{tabular}[c]{@{}c@{}}     @DivyanshMohit @GulBukhari\\ Tum \textcolor{magenta}{apny} Indian ki \textcolor{magenta}{fikkar} Karo\\ Pakistan ko hum	khud \textcolor{magenta}{dykh lyngy}.\\ Mukti \textcolor{magenta}{bahini} 2 nahi ban	 \end{tabular}   &      \begin{tabular}[c]{@{}c@{}} \textcolor{orange}{@BTS\_army\_Fin Also Stade de} \\ \textcolor{orange}{France is preparing for the concert.} \\ \textcolor{orange}{Looks so beautiful! See their} \\ \textcolor{orange}{ post on Instagram https//t.co/OwhP} \end{tabular} &     \begin{tabular}[c]{@{}c@{}} \textcolor{blue}{Now	this i too	much ab}\\ \textcolor{blue}{sare tweet arsal ke support}\\ \textcolor{blue}{me Jab	jiya ka man	nhi}\\	\textcolor{blue}{and wo chai nhi bana}\\ \textcolor{blue}{sakti yasit ke liy}\end{tabular}  \\ \hline

\newcite{barman-etal-2014-code} &      \begin{tabular}[c]{@{}c@{}}
@Liaqat842 tum sahi \textcolor{magenta}{keh	rhy	thy}\\	yeh	zayda buri timings	hain\\ 3 wali match ki subah \textcolor{magenta}{purany}\\	office	bhi	\textcolor{magenta}{jna}	hai	kaam hai\end{tabular} & \begin{tabular}[c]{@{}c@{}}   \textcolor{blue}{ @saadiaafzaal Pagl he	ye	Qaom Jo} \\ \textcolor{blue}{misbah	ka	Cmprezm	imraan	se	kr} \\ \textcolor{blue}{rhe	he. khuda ko	maano kaha} \\ \textcolor{blue}{misbah	kaha imran.. shoib Akhtar}  \end{tabular} &  \begin{tabular}[c]{@{}c@{}}    \textcolor{purple}{@aashikapokharel Haha okay. Office}\\	\textcolor{purple}{time, aba	bus	ma	bore hune wala}\\ \textcolor{purple}{chhu. Also, Alumni	ko imp}\\	\textcolor{purple}{kaam chha. Viber ma aaye hune. :P}\end{tabular}    \\ \hline
\newcite{bohra2018dataset} &     \begin{tabular}[c]{@{}c@{}}Gf khoon peene \textcolor{magenta}{k} liye hoti\\ hai aur apne babu ko \\ \textcolor{magenta}{thana thilane k} liye bas \end{tabular}& \textcolor{blue}{Mere marnay ki ya hate deni ki?}  &   \begin{tabular}[c]{@{}c@{}} \textcolor{purple}{ke karya karta aise hi} \\ \textcolor{purple}{baithe hai.kal ye ghatna}\\ \textcolor{purple}{aap or Hum} \end{tabular} \\ \hline
\newcite{dhar2018enabling} &    \begin{tabular}[c]{@{}c@{}}    Modi ji aap \textcolor{magenta}{jesa} koi \textcolor{magenta}{nhi}\\ \textcolor{magenta}{dhanywad} aap desh ki kitni\\ sewa karte \textcolor{magenta}{hai} \textcolor{magenta}{jese ak} beta\\ apni \textcolor{magenta}{ma} ko poojta \textcolor{magenta}{hai} \end{tabular}    &\textcolor{blue}{Girna samal nai lage}&\begin{tabular}[c]{@{}c@{}}\textcolor{blue}{Etni lambi speech sa kuch}\\ \textcolor{blue}{mi hotta sirf 2 word khna or}\\ \textcolor{blue}{unka suna sa frk atta h........}\\ \textcolor{blue}{sekho i love you sallu??}   \end{tabular}\\ \hline
\newcite{srivastava2020phinc} & \begin{tabular}[c]{@{}c@{}} unhone pehle pic ni \textcolor{magenta}{dkhi ti}\\ kya  \textcolor{magenta}{tmhari} jo milne \textcolor{magenta}{k} baad \\hi ignore \textcolor{magenta}{kia tmhe}...?     \end{tabular}&    \textcolor{orange}{kaun hai ye zaleel insaan?}&\begin{tabular}[c]{@{}c@{}} \textcolor{purple}{@indiantweeter Jain ration}\\ \textcolor{purple}{gap ho jaega.}    \end{tabular}     \\ \hline

\newcite{vijay2018dataset} &   \begin{tabular}[c]{@{}c@{}} 35 sal \textcolor{magenta}{ma} koi hospital \textcolor{magenta}{esa} \\ \textcolor{magenta}{nai} banaya jaha khud\\ ka ilaj \textcolor{magenta}{hosakai}. .. Irony \end{tabular}
& \begin{tabular}[c]{@{}c@{}}\textcolor{blue}{and then the irony,, sab ko jurisakyo} \\ \textcolor{blue}{lahana le kahile juraucha ?} \end{tabular}      &      \begin{tabular}[c]{@{}c@{}}\textcolor{blue}{hi Vanitha Garu  hai Andi this is} \\ \textcolor{blue}{irony , arledy rep icharu ga} \\ \end{tabular}    \      \\ \hline

\newcite{khanuja2020new} &3 kam padey \textcolor{magenta}{they}   & \textcolor{orange}{KASTURI is speaking to his son}      &     \textcolor{purple}{31 minutes time hua} \\ \hline
\end{tabular}}
\caption{Examples from the 10 datasets highlighting the various inherent limitations that could lead to misleading code-mxing metric score. For the marked words in \textcolor{magenta}{\textbf{spelling variations}}, we observe multiple spellings across datasets. We observe that the \textcolor{blue}{\textbf{noisy}} sentences have low \textcolor{blue}{\textbf{readability}}.} 
\label{tab:examples}
\end{table*}

\begin{enumerate}
    \item \textbf{Metric formulation}: Most of the code-mixing metrics are based on the word frequency from different languages in the text. This formulation makes the metric vulnerable to several limitations, such as the bag-of-words model and assigning higher metric scores to meaningless sentences that are difficult to read and comprehend. 
    \item \textbf{Resource limitation}: The existing code-mixed datasets too have several shortcomings, such as noisy and monolingual text (see Table \ref{tab:examples}). Besides, we observe the poor quality of the token-level language identification (LID) systems which are fundamental in calculating the various code-mixing metric scores. 
    \item \textbf{Human annotation}: In the absence of good quality code-mixed LID systems, various works employ human annotators to perform language identification. Evaluating human proficiency is a challenging task since code-mixed languages lacks standard syntax and semantics. Additionally, human annotation is a time and effort extensive process. 
\end{enumerate}

\begin{figure}[!tbh]
    \centering
    \includegraphics[width = 0.50\textwidth]{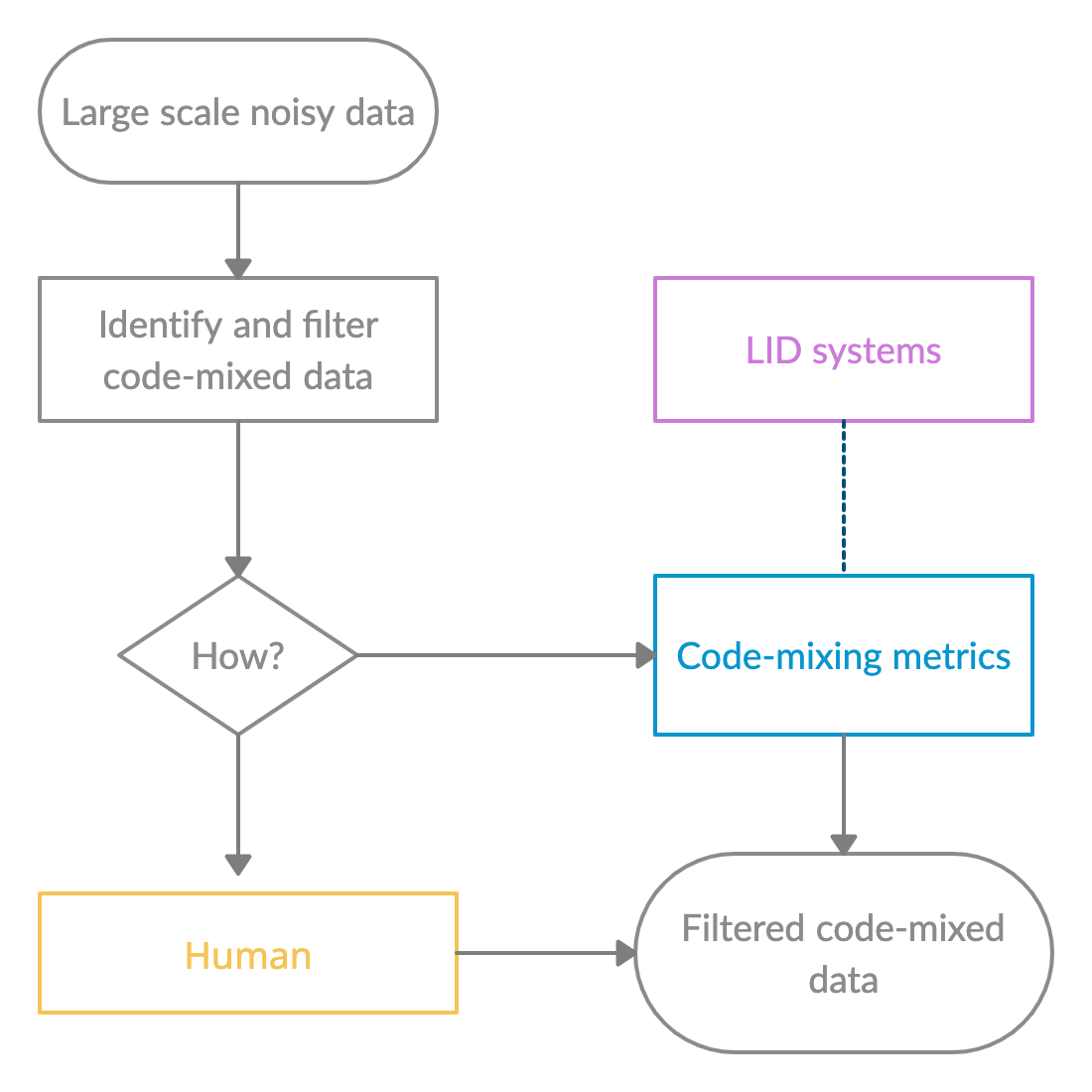}
    \caption{A general flow diagram for identifying and filtering the code-mixed data from the large scale noisy text. We observe three major limitations: \textcolor{blue}{metric formulation}, \textcolor{purple}{resource limitation}, and \textcolor{orange}{human annotation}. There is a time-quality trade-off between the two paths to filter the code-mixed data. Employing humans takes more time and relatively better quality code-mixed sentences as compared to code-mixing metrics that takes less time and shows poor performance.}
    \label{fig:CMI_flow}
\end{figure}

Next, we describe four major limitations that combine one or more than one perspective (see Table \ref{tab:perspective}). Figure \ref{fig:CMI_flow} shows a general flow diagram to obtain the code-mixed data from the large-scale noisy text. It shows the three major bottlenecks (metric formulation, resource limitation, and human annotation) in the entire data filtering process. The resultant code-mixed data is noisy and suffers from several other limitations (see Table \ref{tab:examples}).

\begin{table}[!tbh]
\centering
\small{
\begin{tabular}{|c|c|}
\hline
\textbf{Limitation}     & \textbf{Perspective} \\ \hline
Bag of words      & MF                   \\ \hline
Code-mixed LID     & MF, RL               \\ \hline
Misleading score   & MF, RL, HA           \\ \hline
High inference time & MF, RL, HA           \\ \hline
\end{tabular}}
\caption{Combination of perspectives for each of the limitation to code-mixing metrics. Here, MF: Metric Formulation, RL: Resource Limitation, HA: Human Annotation.}
\label{tab:perspective}
\end{table}

\begin{enumerate}
    \item \textbf{Bag-of-words}: None of the code-mixing metrics consider inherent ordering between the words in the code-mixed sentence\footnote{Note that, \textit{Burstiness} and \textit{Memory} metric only considers span length and not the word ordering within a span.}. This limitation makes these metric scores vulnerable to multiple challenges, such as poor grammatical structure. Figure \ref{fig:bag_of_word} shows examples of good quality code-mixed sentences and corresponding noisy sentences, both having the same metric scores. 
    
\begin{figure}[!tbh]
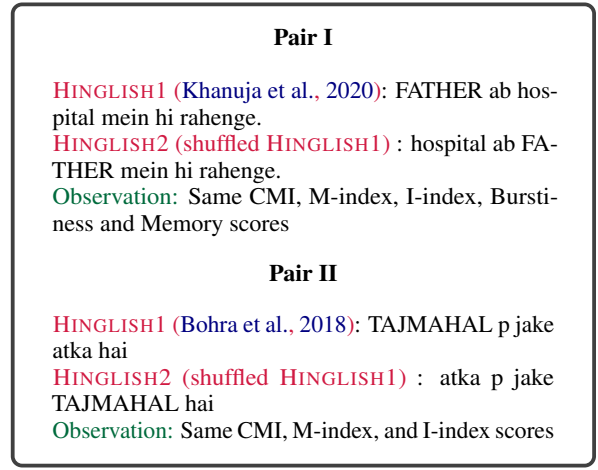

\centering
\small{
\begin{tcolorbox}[colback=white]
\begin{center}
    \textbf{Pair I}
\end{center}
\textcolor{alizarin}{\textsc{Hinglish1} \cite{khanuja2020new}}: FATHER ab hospital mein hi rahenge.  \\
\textcolor{alizarin}{\textsc{Hinglish2} (shuffled \textsc{Hinglish1}) }: hospital ab FATHER mein hi rahenge.  \\
\textcolor{cadmiumgreen}{Observation:} Same CMI, M-index, I-index, Burstiness and Memory scores
\begin{center}
    \textbf{Pair II}
\end{center}
\textcolor{alizarin}{\textsc{Hinglish1} \cite{bohra2018dataset}}: TAJMAHAL p jake atka hai \\
\textcolor{alizarin}{\textsc{Hinglish2} (shuffled \textsc{Hinglish1}) }: atka p jake TAJMAHAL hai  \\
\textcolor{cadmiumgreen}{Observation:} Same CMI, M-index, and I-index scores
\end{tcolorbox}}
\caption{Example to demonstrate the \textit{bag of words} assumption of code-mixing metrics. We shuffle tokens in \textcolor{alizarin}{\textsc{Hinglish1}} to get \textcolor{alizarin}{\textsc{Hinglish2}}. Observation shows that metric scores remain unchanged after the shuffling while the semantic of the original sentence is lost.}
\label{fig:bag_of_word}
\end{figure}

\begin{table*}[!tbh]
\resizebox{\hsize}{!}{
\begin{tabular}{|c|c|c|c|c|c|c|c|c|c|c|c|c|c|c|c|c|c|c|c|c|c|}
\hline
   & @user & bus & office & me & hn & ,    & Sat & thora & thanda & hota & hay & kaam & k  & point & of & view & say & you & know & :)   \\ \hline
\textbf{Langdetect} &
et & id & en & nl & vi & unk & tl & en & en & cs & so & so & sw & fi & en & af & tl & sw & en & unk\\ \hline
\textbf{Polyglot} & en & en & en & en & da & un & en & en & en & to & es & fy & en & en & en & en & en & en & en & un     \\\hline
\textbf{CLD3} & no & la & ja & mi & sv & ja & sd & la & ko & mi & es & et & sl & de & en & en & id & en & en & ja    \\\hline
\textbf{FastText} & en & en & en & en & en & ru & pt & war & en & en & es & az & ja & en & en & en & en & en & en & uz      \\\hline
\textbf{iNLTK} & en & en & en & en & en & en & en & en & en & en & en & en & en & en & en & en & en & en & en & en     \\\hline
\textbf{Human}    & univ& en  & en     & hi & hi & univ & en  & hi    & hi     & hi   & hi  & hi   & hi & en    & en & en   & hi  & en  & en   & univ \\ \hline
\end{tabular}}
\caption{Example to demonstrate the limitations of LID systems in calculating the code-mixing metric scores. Hinglish sentence is from the dataset used in \cite{barman-etal-2014-code}. The language name corresponding to the language code can be found at the corresponding LID system's web page.}
\label{tab:lang_identification}
\end{table*}

\begin{table*}[!tbh]
\resizebox{\hsize}{!}{
\begin{tabular}{|c|c|c|c|c|c|c|c|c|c|c|c|c|c|}
\hline
\textbf{Token}    & @& nehantics & Haan    & yaar & neha & kab  & karega & woh & post  & Usne & na  & sach & mein\\ \hline
\textbf{Language} & O& \textcolor{magenta}{Hin}& Hin     & Hin  & \textcolor{magenta}{Hin}  & Hin  & Hin    & Hin & \textcolor{blue}{Hin}   & Hin  & Hin & Hin  & Hin\\ \hline 
\textbf{Token}    & photoshoot & karna     & chahiye & phir & woh  & post & karega & …   & https & //   & tco & /    & 5RSlSbZNtt \\ \hline
\textbf{Language} & Eng& Hin& Hin     & Hin  & Hin  & \textcolor{blue}{Hin}  & Hin    & O   & Eng   & O    & \textcolor{orange}{Eng} & O    & \textcolor{orange}{Eng}\\ \hline
\end{tabular}}
\centering  (a) Example sentence from \newcite{patwa2020semeval}

\resizebox{\hsize}{!}{
\begin{tabular}{|c|c|c|c|c|c|c|c|c|c|c|c|c|}
\hline
\textbf{Token}    & are     & cricket & se   & sanyas & le   & liya & kya  & viru & aur     & social & service & suru \\ \hline
\textbf{Language} & Hin      & Eng      & Hin   & Hin     & Hin   & Hin   & Hin   & \textcolor{magenta}{Hin}   & Hin      & Eng     & Eng      & Hin   \\ \hline
\textbf{Token}    & kardiya & .& khel & hi     & bhul & gaye & .    & 2    & innings & 0      & n& 0    \\ \hline
\textbf{Language} & Hin      & O    & Hin   & Hin     & Hin   & Hin   & O & O & \textcolor{blue}{Hin}      &  O   & \textcolor{orange}{Hin}      & O \\ \hline
\end{tabular}}
 (b) Example sentence from \newcite{swami2018corpus}
\caption{Example sentences to demonstrate the limitations with the language tags in the current code-mixed datasets. We use the color coding to represent three major reasons for such behaviour: \textcolor{magenta}{ambiguous}, \textcolor{blue}{annotator's proficiency}, and \textcolor{orange}{non-contextual}. `O' in the language tag represent the tag `Other'.}
\label{tab: LID_limitation}
\end{table*}

\item \textbf{Code-mixed language identification}: The presence of more than one language in the code-mixed text presents several challenges for the various downstream NLP tasks such as POS tagging, summarization and named entity recognition. Identifying the token-level language of the code-mixed text is the fundamental step in calculating the code-mixing metric scores. Often various works have employed human annotators to obtain the token-level language tags. However, both human annotators and the language identification systems suffer from the poor token-level language tagging. Table~\ref{tab:lang_identification} shows the variation in the output of five multilingual/code-mixed LID systems (Langdetect\footnote{\url{https://pypi.org/project/langdetect/}}, Polyglot\footnote{\url{https://github.com/aboSamoor/polyglot}}, CLD3\footnote{\url{https://github.com/google/cld3/}}, FastText\footnote{\url{https://fasttext.cc/blog/2017/10/02/blog-post.html}}, and iNLTK\footnote{\url{https://inltk.readthedocs.io/en/latest/index.html}}) on the code-mixed text against human-annotated language tags. Contrasting human-annotated tag sequence, the same metric yields significantly different scores due to variation in the language tag sequence obtained from different LID tools. We identify three major reasons for the poor performance of humans and the LID systems in identifying the language of the code-mixed text: 
\begin{itemize}
\item \textbf{Spelling variations and non-contextual LID}: Spelling variation is one of the most significant challenges in developing code-mixed LID systems. Due to the lack of standard grammar and spellings in code-mixed language, we observe multiple variations of the same word across datasets (see Table \ref{tab:examples}).  For example, Hindi tokens \textit{`hn'} or \textit{`hay'} can also be written as \textit{`hun'} or \textit{`hai'}, respectively. As outlined in Table~\ref{tab:lang_identification}, we observe incorrect language identification by popular multilingual and code-mixed LID systems. This behavior could be highly attributed to the spelling variation of words.  Additionally, the non-contextual language tag sequence generation by LID systems and humans leads to a similar set of challenges (see Table~\ref{tab: LID_limitation}). In both the examples in Table \ref{tab: LID_limitation}, we observe the incorrect language tag to words like \textit{`tco'} and \textit{`n'} due to the missing context by the human annotator. Also, as observed in Table \ref{tab: LID_limitation}, incorrect LID by humans could be attributed to considering the code-mixed tokens out of context.

\item \textbf{Ambiguity}: Ambiguity in identifying named-entities, abbreviations, community-specific jargons, etc., leads to incorrect language identification. Table \ref{tab: LID_limitation} shows the example sentences having incorrect language tags due to ambiguity in the code-mixed sentences. For example, tokens like \textit{`nehatics'}, \textit{`neha'}, and \textit{`viru'} are person named-entities, incorrectly tagged with \textit{hi} tag.

\item \textbf{Annotator's proficiency}: Evaluating the human proficiency for a code-mixed language is much more challenging as compared to the monolingual languages due to lack of standard, dialect variation, and ambiguity in the text. Table \ref{tab: LID_limitation} shows an example of incorrect language annotation by the human annotators, which could be attributed to low human proficiency/varied interpretation of the code-mixed text. For example, English tokens like \textit{`post'} and \textit{`innings'} are tagged as \textit{hi} tokens by human annotators.
\end{itemize}
    
\item \textbf{Misleading score}: We observe several inconsistencies in the interpretation of the code-mixing metric scores. We identify three major reasons for this inconsistent behavior:   
    \begin{itemize}
    \item \textbf{Coherence}: Coherency in a multi-sentence code-mixed text is one of the fundamental properties of good quality data. Future works in code-mixed NLP, such as text summarization, question-answering, and natural language inference, will require highly coherent datasets. However, the current metrics cannot measure the coherency of the code-mixed text. We witness a large number of real scenarios where the code-mixing metric scores for multi-sentence text are high, but the coherency is very poor.  In such cases, the code-mixing metrics in the present form will lead to undesirable behavior. For instance, we query a Hinglish question-answering system \textit{WebShodh}\footnote{\url{http://tts.speech.cs.cmu.edu/webshodh/cmqa.php}} \cite{chandu2017webshodh} with the question: \textit{India ka PM kaun hai? Cricket dekhne jaana hai?} The list of eight probable answers (\textit{`ipl', `puma', `kohli', `sports news feb', `'amazoncom', `sport news nov', `hotstar vip', `rugged flip phone unlocked water shock proof att tmobile metro cricket straight talk consumer cellular carrier cell phones'}) shows the poor performance of the system due to low coherency in the question text (in addition to other architectural limitations) even though the question text is highly code-mixed on various metrics. 
   
    \item \textbf{Readability}: The co-existence of the code-mixed data with the monolingual and the noisy text results in the poor readability of the code-mixed text. The code-mixing metrics do not take into account the readability of the code-mixed text. Low readability of the code-mixed text will also lead to incorrect annotations by the annotators, which will eventually lead to incorrect metric scores for the given data. Table  \ref{tab:examples} shows example sentences from multiple datasets with low readability.
    \item \textbf{Semantics}: The last column in Table~\ref{tab:examples} shows example sentences from multiple datasets where it is extremely difficult to extract the meaning of the code-mixed sentence. Due to the current formulation of the code-mixing metrics where we consider the independent language tokens and the bag-of-words approach, it is not feasible to identify such low semantic sentences.  
\end{itemize}

\item \textbf{High inference time}: We require an efficient automatic NLP system that identifies and filters the code-mixed text from a large-scale noisy text or monolingual text. Even though theoretically, the code-mixing metrics can help identify text with high levels of code-mixing, but practically they fail due to inefficiencies in LID systems. We showcase the inability of LID systems to detect correct language tags (see point 2 above). One possible remedy is to employ humans in language identification. However, human involvement significantly increases the time and the cost of performing the labeling task. Also, human annotations are also prone to errors (see Table \ref{tab: LID_limitation}).  We might also need task-specific annotations (e.g., POS tags, NER, etc.) which will further increase the time and cost of the annotation task. Due to this reason, we see majority of the datasets (see Table~\ref{tab:datasets}) relatively smaller in size (<5000 data points). Human annotation significantly increases the inference time in calculating the code-mixing metric scores.
\end{enumerate}

\section{Conclusion and Future Work}
\label{sec:conclusion}
In this paper, we extensively discuss the limitations of code-mixing metrics. We explored 10 Hinglish datasets for presenting examples to support our claims. Overall, we showcase the need for extensive efforts in addressing these limitations. In the future, we plan to develop a robust code-mixing metric that measures the extent of code-mixing and quantifies the readability and grammatical correctness of the text. Also, we aim to create a large-scale Hinglish dataset with manual token-level language annotation.

\bibliographystyle{acl_natbib}
\bibliography{naacl2021}
\flushend
\end{document}